# Design Strategies for the Geometric Synthesis of Orthoglide-type Mechanisms

Pashkevich A.[1], Wenger P.[2] and Chablat D.[2*]

[1]Belarusian State University of Informatics and Radioelectronics
P. Brovka str, 220027 Minsk, Republic of Belarus
pap@bsuir.unibel.by
Tel. 375 17 239 86 73, Fax. 375 17 231 09 14

[2]Institut de Recherche en Communications et Cybernétique de Nantes[1]
1, rue de la Noë B.P. 6597, 44321 Nantes Cedex 3, France
{Philippe.Wenger, Damien.Chablat}@irccyn.ec-nantes.fr
Tel. 33 2 40 37 69 54, Fax. 33 2 40 37 69 30

## ABSTRACT

The paper addresses the geometric synthesis of Orthoglide-type mechanism, a family of 3-DOF parallel manipulators for rapid machining applications, which combine advantages of both serial mechanisms and parallel kinematic architectures. These manipulators possess quasi-isotropic kinematic performances and are made up of three actuated fixed prismatic joints, which are mutually orthogonal and connected to a mobile platform via three parallelogram chains. The platform moves in the Cartesian space with fixed orientation, similar to conventional XYZ-machine. Three strategies have been proposed to define the Orthoglide geometric parameters (manipulator link lengths and actuated joint limits) as functions of a cubic workspace size and dextrous properties expressed by bounds on the velocity transmission factors, manipulability or the Jacobian condition number. Low inertia and intrinsic stiffness have been set as additional design goals expressed by the minimal link length requirement. For each design strategy, analytical expressions for computing the Orthoglide parameters are proposed. It is showed that the proposed strategies yield Pareto-optimal solutions, which differ by the kinematic performances outside the prescribed Cartesian cube (but within the workspace bounded by the actuated joint limits). The proposed technique is illustrated with numerical examples for the Orthoglide prototype design.

---

# 1.   INTRODUCTION

Parallel kinematic machines (PKM) are commonly claimed to offer several advantages over their serial counterparts, such as high structural rigidity, better payload-to-weight ratio, high dynamic capacities and high accuracy [1-3]. Thus, they are prudently considered as promising alternatives for high-speed machining and have gained essential attention of a number of companies and researchers. Since the first prototype presented in 1994 during the IMTS in Chicago by Gidding&Lewis (the VARIAX), many other parallel manipulators have appeared. However, most of the existing PKM still suffer from two major drawbacks, namely, a complex workspace and highly non-linear input/output relations [4, 5].

For most PKM, the Jacobian matrix, which relates the joint rates to the output velocities, is not isotropic. Consequently, the performances (e.g. maximum speeds, forces, accuracy and rigidity) vary considerably for different points in the Cartesian workspace and for different directions at one given point. This is a serious disadvantage for machining applications [6, 7], which require regular workspace shape and acceptable kinetostatic performances throughout. In milling applications, for instance, the machining conditions must remain constant along the whole tool path [8]. Nevertheless, in many research papers, this criterion is not taken into account in the algorithmic methods used for the optimization of the workspace volume [9, 10].

In contrast, for the conventional XYZ-machines, the tool motion in any direction is linearly related to the motions of the actuated axes. Also, the performances are constant throughout the Cartesian parallelepiped workspace. The only drawback is inherent to the serial arrangement of the links, which causes poor dynamic performances. So, in recent years, several new parallel kinematic structures have been proposed. In particular, a 3-dof translational mechanism with gliding foot points was found in three separate works to be fully isotropic throughout the Cartesian workspace [11-13]. Although this manipulator behaves like the conventional Cartesian mechanism, its legs are rather bulky to assure stiffness. The latter motivates further research in PKM architecture that seeks for compromise solutions, which admit a partial isotropy in favour of other manipulator features.

One of such compromise solutions is the Orthoglide proposed by Wenger and Chablat [14], which was derived from a Delta-type architecture with three fixed linear joints and three articulated parallelograms. As follows from the previous works, this manipulator possesses good (almost isotropic) kinetostatic performances and also has some technological advantages, such as (i) symmetrical design; (ii) quasi-isotropic workspace; and (iii) low inertia effects [15]. In a previous work, the Orthoglide was optimised with respect to the Jacobian matrix conditioning and transmission factor limits throughout a prescribed Cartesian workspace [16].

This paper further contributes to the Orthoglide kinematic synthesis and focuses on the comparison of different design strategies and inherited criteria. It proposes a systematic design procedure to define the manipulator geometric parameters (the actuated joint limits and the link lengths) as function of the prescribed cubic workspace size and performances measure bounds. The reminder of the paper is organized as follows. Section 2 briefly describes the Orthoglide kinematics and defines the design goals. Section 3 investigates the manipulator performances through the workspace. Section 4 deals with the design of the dextrous workspace with bounded manipulability, condition number and velocity transmission factors. Section 5 focuses on defining the largest cube inscribed in the dextrous workspace. Section 6 illustrates the proposed design strategies by numerical examples and also contains some discussions. And, finally, Section 7 summarises the main contributions of the paper.

## 2. ORTHOGLIDE KINEMATICS AND DESIGN GOALS

### 2.1. MANIPULATOR GEOMETRY

The kinematic architecture of the Orthoglide is shown in Fig. 1. It consists of three identical parallel chains that may be formally described as $PRP_aR$, where $P$, $R$ and $P_a$ denote the prismatic, revolute, and parallelogram joints respectively. The mechanism input is made up of three actuated orthogonal prismatic joints. The output machinery (with a tool mounting flange) is connected to the prismatic joints through a set of three parallelograms, so that it is restricted for translational movements only.

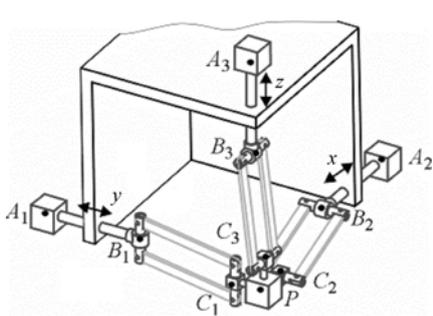

Fig. 1. Kinematic architecture of the Orthoglide

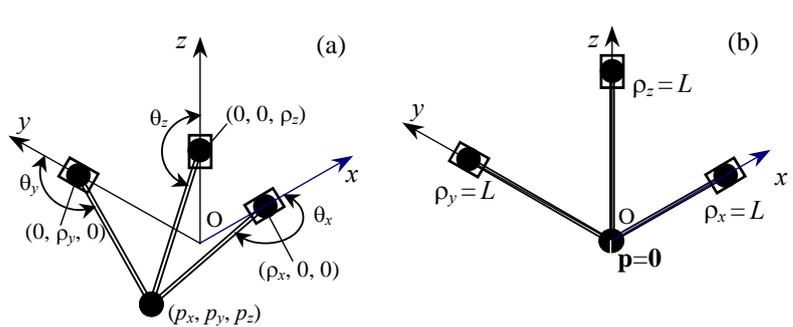

Fig. 2. Orthoglide simplified model (a) and its "zero" configuration (b).

Because of its symmetrical structure, the Orthoglide can be presented in a simplified model, which consists of three bar links connected by spherical joints to the tool centre point at one side and to the corresponding prismatic joints at another side (Fig. 2a).

Thus, if the origin of a reference frame is located at the intersection of the prismatic joint axes and the $x, y, z$-axes are directed along them, the manipulator geometry may be described by the equations

$$\left( p_x - \rho_x \right)^2 + p_y^2 + p_z^2 = L^2; \qquad p_x^2 + \left( p_y - \rho_y \right)^2 + p_z^2 = L^2; \qquad p_x^2 + p_y^2 + \left( p_z - \rho_z \right)^2 = L^2 \qquad (1)$$

where $\mathbf{p} = \left( p_x, p_y, p_z \right)$ is the output position vector, $\rho = \left( \rho_x, \rho_y, \rho_z \right)$ is the input vector of the prismatic joints variables, and $L$ is the length of the parallelogram principal links. It should be noted that, for this convention, the "zero" position $\mathbf{p}_0 = \left( 0, 0, 0 \right)$ corresponds to the joints variables $\rho_0 = \left( L, L, L \right)$, see Fig. 2b.

It is also worth mentioning that the Orthoglide geometry and relevant manufacturing technology impose the following constraints on the joint variables

$$0 < \rho_x \le 2L \; ; \qquad 0 < \rho_y \le 2L \; ; \quad 0 < \rho_z \le 2L, \qquad (2)$$

which essentially influence on the workspace shape. While the upper bound is implicit and obvious, the lower one is caused by practical reasons, since safe mechanical design encourages avoiding risk of simultaneous location of prismatic joints in the same point of the Cartesian workspace. Hence the kinematic synthesis must produce required joint limits within (2).

### 2.2. INVERSE KINEMATICS

From equations (1), the inverse kinematic relations can be derived in a straightforward way

$$\rho_x = p_x + s_x \sqrt{L^2 - p_y^2 - p_z^2} \; ; \qquad \rho_y = p_y + s_y \sqrt{L^2 - p_x^2 - p_z^2} \; ; \qquad \rho_z = p_z + s_z \sqrt{L^2 - p_x^2 - p_y^2} \qquad (3)$$

where $s_x, s_y, s_z \in \{\pm 1\}$ are the configuration indices defined as signs of $\rho_x - p_x$, $\rho_y - p_y$, $\rho_z - p_z$ respectively. Their geometrical meaning is illustrated by Fig. 2a, where $\theta_x, \theta_y, \theta_z$ are the angles between the bar links and corresponding prismatic joint axes. It can be easily proved that $s = +1$ if $\theta_a \in (90^\circ, 180^\circ)$ and $s = -1$ if $\theta_a \in (0^\circ, 90^\circ)$, where the subscript $a$ belongs to the set $a \in \{x, y, z\}$. It should be also stressed that the border ($\theta = 90^\circ$) corresponds to the serial singularity (when the link is orthogonal to the relevant translational axis and the input joint motion does not produce the end-point displacement), so corresponding Cartesian points must be excluded from the Orthoglide workspace during the design.

It is obvious that expressions (3) define eight different solutions to the inverse kinematics and their existence requires the workspace points to belong to a volume bounded by the intersection of three cylinders $C_L = \left\{ \mathbf{p} \mid p_x^2 + p_y^2 \le L^2; \; p_x^2 + p_z^2 \le L^2; \; p_y^2 + p_z^2 \le L^2 \right\}$. However, the joint limits (2) impose additional constraints, which reduce a potential solution set. For example, for the "zero" location $\mathbf{p}_0 = \left( 0, 0, 0 \right)$, the equations (3) give eight solutions $\rho = \left( \pm L, \pm L, \pm L \right)$ but only one of them is feasible. As proved in [17], with respect to number of inverse kinematic solutions, the Orthoglide with joint limits (2) admits only 2 alternatives: (i) a single inverse kinematic solution $(s_x, s_y, s_z = +1)$ inside the sphere $S_L = \left\{ \mathbf{p} \in C_L \middle| \ p_x^2 + p_y^2 + p_z^2 < L^2 \right\}$; and (ii) eight inverse kinematic solutions $(s_x, s_y, s_z \in \{\pm 1\})$ inside $G_L = \left\{ \mathbf{p} \in C_L \mid p_x, p_y, p_z > 0; \ \ p_x^2 + p_y^2 + p_z^2 > L^2 \right\}$. It can be also proved that the border between these two cases corresponds to the serial singularity. Hence, the kinematic synthesis must focus on the location of the workspace inside of the sphere $S_L$.

## 2.3. DIRECT KINEMATICS

After subtracting three possible pairs of the equations (1) and analysis of the differences, the Cartesian coordinates $p_x$, $p_y$, $p_z$ can be expressed as

$$p_x = \frac{\rho_x}{2} + \frac{t}{\rho_x}; \ \ p_y = \frac{\rho_y}{2} + \frac{t}{\rho_y}; \ \ p_z = \frac{\rho_z}{2} + \frac{t}{\rho_z}, \tag{4}$$

where $t$ is an auxiliary scalar variable. This reduces the direct kinematics to the solution of a quadratic equation, $At^2 + Bt + C = 0$ with coefficients $A = \left( \rho_x \rho_y \right)^2 + \left( \rho_x \rho_z \right)^2 + \left( \rho_y \rho_z \right)^2$; $B = \left( \rho_x \rho_y \rho_z \right)^2$; $C = \left( \rho_x^2 + \rho_y^2 + \rho_z^2 - 4L^2 \right)\left( \rho_x \rho_y \rho_z \right)^2 \big/ 4$. The quadratic formula yields two solutions $t = (-B + m\sqrt{B^2 - 4AC}) \big/ (2A)$ that differ by the configuration index $m = \pm 1$, which, from a geometrical point of view, distinguishes two possible locations of the target point with respect to the plane passing through the prismatic joint centres. Algebraically, this index can be defined as $m = \mathrm{sgn}\left( p_x \rho_x^{-1} + p_y \rho_y^{-1} + p_z \rho_z^{-1} - 1 \right)$. It should be stressed that the case $B^2 = 4AC$ corresponds to a parallel singularity, so corresponding joint coordinates must be excluded during the design.

It is obvious that the direct kinematic solution exists if and only if $B^2 \geq 4AC$, which defines a closed region in the joint variable space $\Re_L = \left\{ \mathbf{\rho} \mid \left( \rho_x^2 + \rho_y^2 + \rho_z^2 - 4L^2 \right)\left( \rho_x^{-2} + \rho_y^{-2} + \rho_z^{-2} \right) \leq 1 \right\}$. Taking into account (2), the feasible joint space may be presented as $\Re_L^+ = \left\{ \mathbf{\rho} \in \Re_L \mid \rho_x, \rho_y, \rho_z \geq 0 \ \right\}$. Hence, with respect to the number of direct kinematic solutions, the Orthoglide with joint limits (2) admits 2 alternatives [17]: (i) two direct kinematic solutions $(m = \pm 1)$ inside the region $\Re_L^+$; and (ii) a single direct kinematic solution on the positive border of the region $\Re_L^+$. Since the second case corresponds to the singularity, the kinematic synthesis must focus on using the inner part of $\Re_L^+$.

## 2.4. DESIGN GOALS AND PARAMETERS

Because the Orthoglide is dedicated to general 3-axis machining, its kinematic performances should be close to the performances of the classical XYZ-machine. Therefore, the design goals may be stated as follows:

(i) manipulator workspace should be close to a cube of prescribed size;

(ii) kinematic performances within this cube should be quasi-isotropic;

(iii) link lengths should be minimal to lower the manufacturing costs.

The requirements (i) and (ii) will be satisfied in Section 4 by constraining the manipulability, condition number and/or velocity transmission factor inside the Cartesian workspace bounded by the joint limits. To fulfil requirement (iii), Section 5 evaluates the largest cube inscribed in this workspace, which defines the smallest link lengths required to achieve the prescribed cube size.

The design parameters to be optimised are the parallelogram length $L$, the actuated joint limits $(\rho_{min}, \rho_{max})$ and related location of the prescribed cube $(p_{min}, p_{max})$. Taking into account the linear relation between $L$ and the workspace size, the design process is decomposed into two stages:

(i)   defining the joint limits $(\rho_{min}, \rho_{max})$ and the largest cube size/location $(p_{min}, p_{max})$
       to satisfy given kinematic performances for the normalised manipulator ($L = 1$);

(ii)  scaling the normalised manipulator parameters to achieve the prescribed size of the cubic
       workspace.

Numerical example for this two-stage design process is given in Section 6.

## 3.   JACOBIAN ANALYSIS

### 3.1. JACOBIAN MATRIX

As follows from the previous Section and a companion paper [17], the singularity-free workspace of the Orthoglide $W_0$ is located within the sphere $S_L$ of radius $L$ with the centre point $(0, 0, 0)$ and bounded by the parallel "*flat*" singularity surface in the first octant (Fig. 3). Also, the remaining part of the sphere surface corresponds to the parallel "*bar*" singularity. Hence, the kinematic design should define the inner part of this workspace that possesses the desired kinematic properties.

Mathematically, these properties are defined by the manipulator Jacobian describing the differential mapping from the jointspace to the workspace (or vice versa). For the Orthoglide, it is more convenient to express analytically the inverse Jacobian, which is derived from (1) in a straightforward way:

$$\mathbf{J}^{-1}(\mathbf{p}, \boldsymbol{\rho}) = \begin{bmatrix} 1 & p_y/(p_x\text{-}\rho_x) & p_z/(p_x\text{-}\rho_x) \\ p_x/(p_y\text{-}\rho_y) & 1 & p_z/(p_y\text{-}\rho_y) \\ p_x/(p_z\text{-}\rho_z) & p_y/(p_z\text{-}\rho_z) & 1 \end{bmatrix} \tag{5}$$

Accordingly, the determinant of the Jacobian may be expressed as

$$\det\left(\mathbf{J}^{-1}\right) = \frac{p_x\rho_y\rho_z + \rho_x p_y\rho_z + \rho_x\rho_y p_z - \rho_x\rho_y\rho_z}{\left(p_x - \rho_x\right)\left(p_y - \rho_y\right)\left(p_z - \rho_z\right)} \tag{6}$$

and admits two cases of ill-conditioning, $\det(\mathbf{J}) = 0$ and $\det(\mathbf{J}^{-1}) = 0$, corresponding to the serial and parallel singularities mentioned above. It is also clear that the full isotropy is achieved only in the "zero" point $\mathbf{p}_0 = (0, 0, 0)$, where the Jacobian reduces to the identity matrix: $\mathbf{J}_0 = \mathbf{I}$.

### 3.2. Q-AXIS PROPERTIES

Since the Orthoglide workspace is symmetrical with respect to the axes *x, y, z*, its kinematic design requires a detailed study of the points belonging to the Q-axis, which is the bisector line of the first octant [16]. For this axis, let us denote $p_x = p_y = p_z = p$ and, consequently, $\rho_x = \rho_y = \rho_z = \rho$. Then, as follows from (5), the inverse Jacobian may be presented as

$$\mathbf{J}^{-1}(\chi) = \begin{bmatrix} 1 & \chi & \chi \\ \chi & 1 & \chi \\ \chi & \chi & 1 \end{bmatrix} \qquad (7)$$

where $\chi$ is the dimensionless scalar parameter expressed as $\chi = -p \big/ \sqrt{L^2 - 2p^2}$ and related to the input/output variables via the expressions $p = -\chi L \big/ \sqrt{1 + 2\chi^2}$; $\rho = (1-\chi)L \big/ \sqrt{1 + 2\chi^2}$. To define the feasible range of the parameter $\chi$, let us consider specific points belonging to the Q-axis (see Fig. 4 and Table 1). They include three parallel singularity points $P_1$, $P_2$, $P_3$ and one serial singularity point $P_4$. As follows from the analysis, the singularity-free region of the Q-axis is bounded by the interval $\chi \in (-0.5, 1.0)$, which corresponds to the coordinate ranges $p \in \left( -L/\sqrt{3}, L/\sqrt{6} \right)$; $\rho \in \left( 0, \sqrt{3/2}\,L \right)$. It is important for the kinematic design that, within these limits, the relation between the coordinates *p, $\rho$* and the parameter $\chi$ is monotonously decreasing (see Table 1). It should be also noted that the employed parameterisation may be converted to the one used in [16] by defining $\chi = -\tan(\theta)$, where $\theta$ is the angle between the manipulator links and corresponding prismatic joint axes.

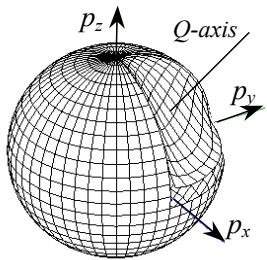

Fig. 3. The singularity-free workspace of the Orthoglide (97.2% of the sphere volume)

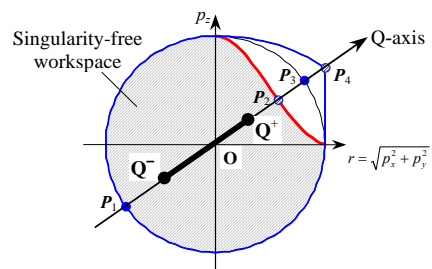

Fig 4. Workspace regions for the Q-axis

Table 1. Specific points in the Q-axis for the unit manipulator (*L*=1)

| Feature | $P_1$ | O | $P_2$ | $P_3$ | $P_4$ |
|---------|-------|---|-------|-------|-------|
| $p$ | $-\sqrt{1/3}$ | 0 | $\sqrt{1/6}$ | $\sqrt{1/3}$ | $\sqrt{1/2}$ |
| $\rho$ | 0 | 1 | $\sqrt{3/2}$ | $\sqrt{4/3}$ | $\sqrt{1/2}$ |
| $\chi$ | 1 | 0 | -0.5 | -1 | $-\infty$ |
| $\det(\mathbf{J})$ | $\infty$ | 1 | $\infty$ | $\infty$ | 0 |

## 4.   DEXTERITY-BASED DESIGN

Since the design specifications require the manipulator to possess the quasi-isotropic kinematics [18-20], the original joint limits (2) must be narrowed to increase the distance from the dextrous workspace points to the singularities. In this section, the desired joint limits are computed using the Q-axis technique, which reduces the problem to locating two points $Q^+$ and $Q^-$ on the bisector line (see Fig. 4). These points bound the Q-axis region with the required properties and, therefore, define the joint limits. It is obvious that the interval $[Q^+, Q^-]$ must include the fully-isotropic "zero" point O, and the kinematic performances at the points $Q^+$, $Q^-$ should be similar. To compute the joint limits, we apply three different criteria evaluating the workspace dexterity. It should be also mentioned that all results of this Section are valid for the "unit" manipulator (*L*=1), which will be scaled on the subsequent design steps.

### 4.1.   CONSTRAINING THE MANIPULABILITY

The manipulator manipulability $w = \sqrt{\det\left(\mathbf{J}^{-1}\mathbf{J}^{-T}\right)}$ is the simplest performance measure assessing the dexterity [21], which is the product of the singular values of the Jacobian or its inverse. For the Q-axis, where $\mathbf{J}^{-1}$ is a square and symmetrical matrix, the manipulability can be computed as

$$w = \left|\det\left(\mathbf{J}^{-1}\right)\right| = \left(1-\chi\right)^2 \cdot \left|1+2\chi\right| \qquad (8)$$

where $\chi \in \,]-0.5, 1.0[$ . As follows from (6), the maximum value of the manipulability $w$ is equal to 1 and is achieved in the "zero" (isotropic) point:

$$\det\left(\mathbf{J}^{-1}(\mathbf{p}_0)\right) = 1; \qquad \det\left(\mathbf{J}^{-1}(\mathbf{p})\right) < 1 \;\; if \;\; \mathbf{p} \neq \mathbf{p}_0 \qquad (9)$$

Therefore, the joint limits can be found from the inequality

$$\det\left(\mathbf{J}^{-1}(\rho)\right) \geq \Delta \qquad \forall \rho \in [\rho_{\min,} \;\; \rho_{\max}] \; , \qquad (10)$$

where $\Delta$ is the prescribed lower bound of the manipulability ($\Delta < 1$). Since the relation $\rho(\chi)$ is monotonous and $\chi = 0$ corresponds to the isotropic posture (see sub-section 3.2), the desired parameter range can be obtained from the cubic equation $2\chi^3 - 3\chi^2 + (1-\Delta) = 0$ by selecting two roots closest to zero. Applying the trigonometric method, it can be obtained that $\chi_1 = 0.5 - \cos(\varphi/3)$, $\quad \chi_2 = 0.5 + \cos(\varphi/3 - \pi/3)$, $\quad$ where $\varphi = a\cos(1 - 2\Delta)$, $\quad \chi_1 < 0$, $\quad \chi_2 > 0$, and, consequently, $\quad \rho_{\min} = \rho(\chi_2)$; $\quad \rho_{\max} = \rho(\chi_1)$ and $\quad p_{\min} = p(\chi_2)$; $\quad p_{\max} = p(\chi_1)$, where functions $\rho(\chi)$ and $p(\chi)$ are defined in sub-section 3.2. The graphical interpretation of this result is presented in Fig. 5. The open question, however, is how to interpret the manipulability design specification $\Delta$ in engineering sense, to be understandable for the designer with a practical background.

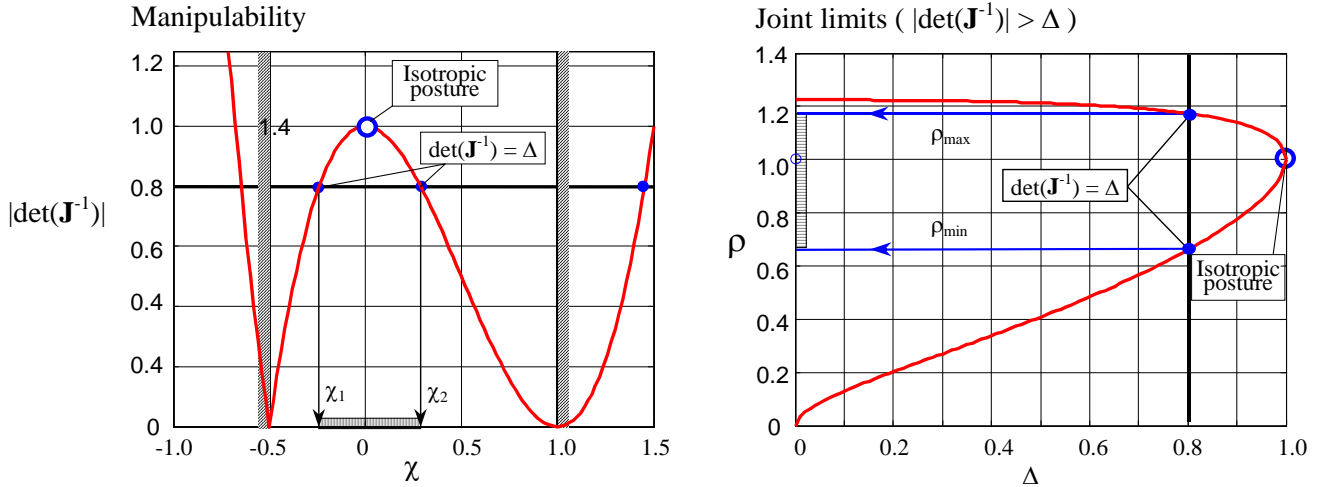

Fig. 5. Computing joint limits via the manipulability index.

## 4.2. CONSTRAINING THE CONDITION NUMBER

The Jacobian condition number evaluates the distance to the singularities by the ratio of the largest to the smallest matrix eigenvalues, which is also the ratio of the largest and smallest axis length of the manipulability ellipsoid [21]. As follows from (5), the Orthoglide condition number achieves its best value (equal to 1) in the zero point, while in other workspace points it is greater than 1:

$$\text{cond}\left(\mathbf{J}^{-1}(\mathbf{p}_0)\right) = 1 \qquad \text{cond}\left(\mathbf{J}^{-1}(\mathbf{p})\right) > 1 \quad if \quad \mathbf{p} \neq \mathbf{p}_0 \tag{11}$$

Hence, the joint limits can be found from the inequality

$$\text{cond}\left(\mathbf{J}^{-1}(\rho)\right) \leq \delta \quad \forall \rho \in [\rho_{\min}, \ \rho_{\max}], \tag{12}$$

where $\delta$ is the admitted upper bound of this performance index ($\delta > 1$). Since along the Q-axis the inverse Jacobian is symmetrical, the condition number can be computed via the ratio of the largest to the smallest eigenvalues of $\mathbf{J}^{-1}$. The relevant characteristic equation $\det(\mathbf{J}^{-1} - \lambda \mathbf{I}) = 0$ may be rewritten as $\lambda^3 - 3\lambda^2 + 3(1 - \chi^2)\lambda - (1 - \chi)^2(2\chi + 1) = 0$. Its analytical solution yields $\lambda_1 = 1 + 2\chi$; $\quad \lambda_{2,3} = 1 - \chi$. Therefore, the condition number for the Q-axis can be expressed as

$$\text{cond}\left(\mathbf{J}^{-1}(\chi)\right) = \begin{cases} 1 + 3\chi/(1-\chi) & if \quad \chi \in ]0, \ 1[ \\ 1 - 3\chi/(1+2\chi) & if \quad \chi \in ]\text{-}0.5, \ 0] \end{cases} \tag{13}$$

and the desired parameter range $[\chi_1, \chi_2]$ can be obtained from the equations $\chi_1 = -(\delta-1)/(2\delta+1)$, $\chi_2 = (\delta-1)/(\delta+2)$. Hence, $\rho_{\min} = \rho(\chi_2); \quad \rho_{\max} = \rho(\chi_1)$ and $p_{\min} = p(\chi_2); \quad p_{\max} = p(\chi_1)$, where functions $\rho(\chi)$ and $p(\chi)$ are defined in sub-section 3.2. The graphical interpretation of this result is presented in Fig. 6. The question of defining a reasonable value of $\delta$ is simpler is this case because it possesses clearer geometric meaning and is rather understandable for practising engineers.

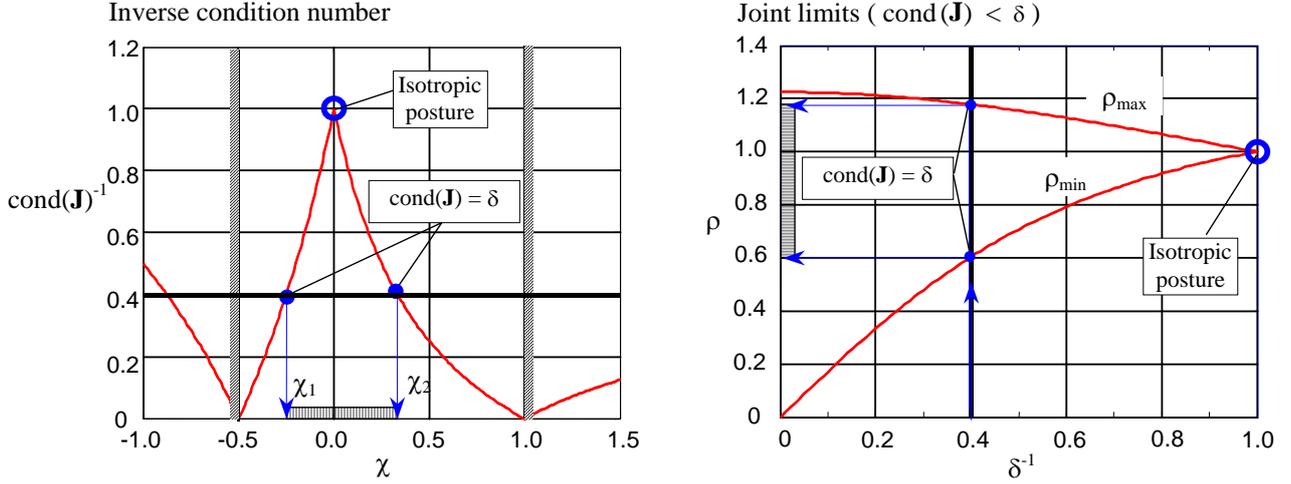

Fig. 6. Computing joint limits via the condition number.

### 4.3. CONSTRAINING THE VELOCITY TRANSMISSION FACTOR

The velocity transmission factor assesses the ratio of the manipulator end-point velocity and velocity of the corresponding point in the joint space. For a given workspace point $\mathbf{p}$ and direction of motion $\mathbf{e}$, it can be computed via the Jacobian as $\lambda(\mathbf{p},\mathbf{e}) = ||\mathbf{J}^{-1}(\mathbf{p})\cdot\mathbf{e}||^{-1}$ where $\mathbf{e}^T\mathbf{e} = 1$. As known from the matrix theory, the deviation of this factor for the fixed $\mathbf{p}$ is bounded by the smallest and largest singular values of $\mathbf{J}$. Geometrically, this performance index is directly related to the manipulability ellipsoid, which in the previous Sub-Section was evaluated by the ratio of its longest and shortest axes, while here these axes are assessed separately.

As follows from (5), the Orthoglide velocity transmission factor does not depend on the direction of motion in the zero point and in the remaining points it varies depending on $\mathbf{e}$:

$$\min_{\mathbf{e}} \lambda(\mathbf{p},\mathbf{e}) < 1; \quad \max_{\mathbf{e}} \lambda(\mathbf{p},\mathbf{e}) > 1 \quad \textit{if} \quad \mathbf{p} \neq \mathbf{p}_0 \qquad (14)$$

Hence, for this performance measure, the joint limits can be found from the inequality

$$\lambda_{\min} \leq \lambda(\rho,\mathbf{e}) \leq \lambda_{\max} \qquad \forall \rho \in [\rho_{\min}, \ \rho_{\max}], \quad \forall \mathbf{e}: \ \|\mathbf{e}\| = 1, \qquad (15)$$

where $\lambda(\rho,\mathbf{e})$ denotes the velocity transmission factor along the Q-axis, and $\lambda_{\min}$, $\lambda_{\max}$ are the design specifications ($\lambda_{\min} < 1 < \lambda_{\max}$).

Since along the Q-axis the Jacobian is symmetrical, the transmission factor range may be derived from the eigenvalues obtained in the Sub-Section 4.2. It has been also proved that the eigenvalue $\lambda_1 = 1 + 2\chi$ corresponds to the eigenvector directed along the Q-axis, and two remaining coinciding eigenvalues $\lambda_{2,3} = 1 - \chi$ correspond to the eigenvectors, which are perpendicular to this axis. So, the desired parameter range $[\chi_1, \chi_2]$ can be computed from the expressions

$$\chi_1 = \max\left\{1 - \lambda_{\max}, \, (\lambda_{\min} - 1)/2\right\}; \quad \chi_2 = \min\left\{1 - \lambda_{\min}, \, (\lambda_{\max} - 1)/2\right\} \qquad (16)$$

Graphical interpretation of this result is presented in Fig. 7. The question of defining reasonable values of $\lambda_{\min}$, $\lambda_{\max}$ is very clear in this case. For instance $\lambda \in [\mu, 1/\mu]$ with $\mu \in [0.5, \, 1.0]$ possesses sensible meaning and is quite understandable for practising engineers. Impact of the transmission factor bounding on the dextrous workspace shape/size is also illustrated in Table 2, where all cases are quantified relative to the volume of the singularity-free workspace $V_0$ (see Fig. 3). To generate these shapes, we executed spanning of all possible directions from the isotropic point and dichotomic search for the line segments satisfying the kinematic constraints.

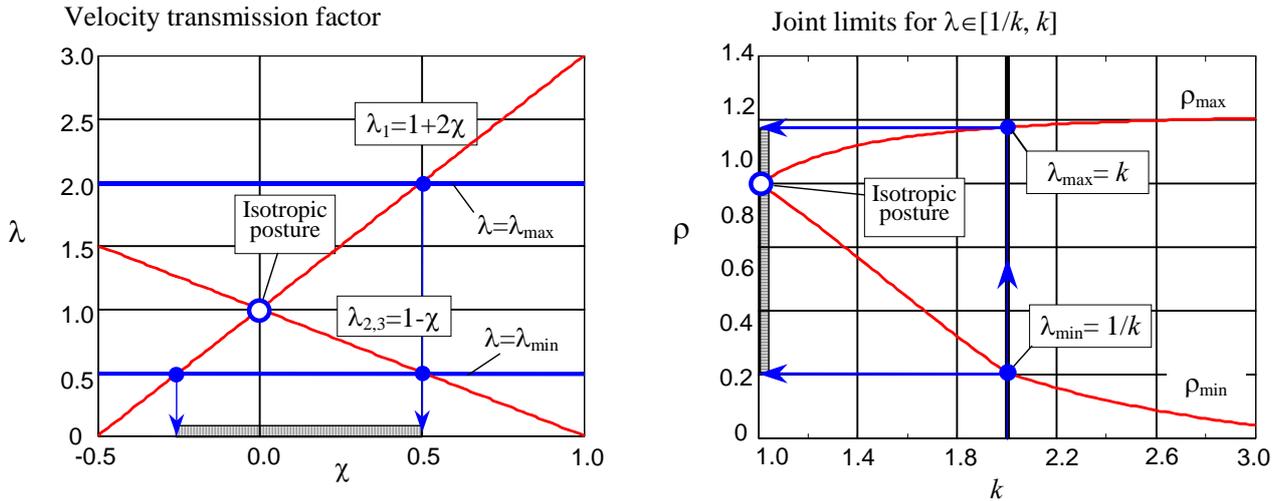

**Fig. 7.** Computing the joint limits the velocity transmission factors.

Table 2
Dextrous workspace for different bounds on the velocity transmission factor

| Lower bounding | Two-sided bounding | Upper bounding |
|---|---|---|
| $\mu \geq 1/3$ | $\mu \in [1/3, \, 3]$ | $\mu \leq 3$ |
| $0.84 \cdot V_0$ | $0.67 \cdot V_0$ | $0.72 \cdot V_0$ |

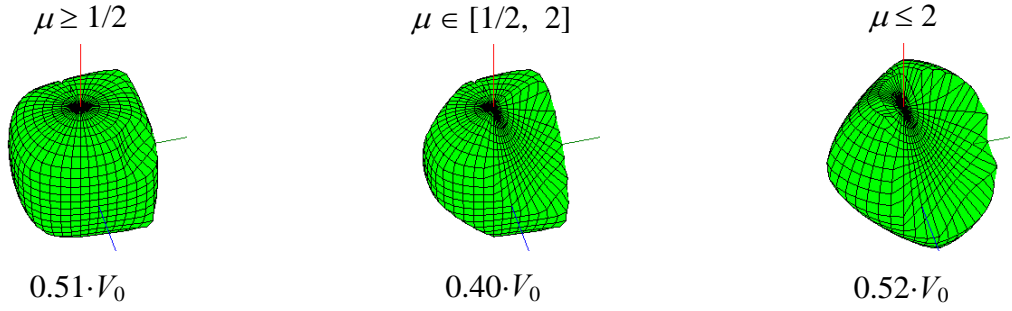

$\mu \geq 1/2$          $\mu \in [1/2, \; 2]$          $\mu \leq 2$

$0.51 \cdot V_0$          $0.40 \cdot V_0$          $0.52 \cdot V_0$

## 5.   WORKSPACE-BASED DESIGN

After applying the Q-axis technique, which yields the joint limits ensuring the prescribed dexterity for the bisector line, the whole workspace must be verified for kinematic performances. In this Section, this problem is solved by identifying and evaluating the workspace "critical points" and relevant definition of the joint limits. Then, the largest cube is inscribed in the dextrous workspace of the unit manipulator, which gives the scaling factor to meet the specifications for the desired cubic workspace size. It should be noted that here the manipulator dexterity is evaluated by the velocity transmission factors, which, as stated above, have advantages over the manipulability and condition number indices in practical applications. But the main results are also generalised for the manipulability and condition number criterions.

### 5.1.   WORKSPACE CRITICAL POINTS

Let us consider first the unit Orthoglide ($L$=1) with given joint limits $\rho_{\min} \in [0, 1]$, $\rho_{\max} \in \left[1, \sqrt{3/2}\right]$ and estimate the velocity transmission factors $\mu_{\min}$, $\mu_{\max}$ over the corresponding workspace, which is bounded by six surfaces presented in Fig. 8. Since the manipulator has a symmetric geometric structure, the candidate points for extreme values of $\mu$ are located symmetrically on the workspace boundary and must be selected from the following sets:

(i) *vertex points,* for which all 3 joint coordinates $\rho_x$, $\rho_y$, $\rho_z$ are equal to either $\rho_{\min}$ or $\rho_{\max}$;

(ii) *edge points,* for which 2 of 3 joint coordinates $\rho_x$, $\rho_y$, $\rho_z$ are equal to either $\rho_{\min}$ or $\rho_{\max}$;

(iii) *face points*, for which 1 of 3 joint coordinates $\rho_x$, $\rho_y$, $\rho_z$ is equal to either $\rho_{\min}$ or $\rho_{\max}$;

It is also obvious that the inner workspace points possess better dexterity than their boundary counterparts (since the straight line motion from the zero point to any boundary point causes monotonous changing of the angles $\theta_x, \theta_y, \theta_z$ and corresponding decreasing of the transmission factors for each axis). Besides, as follows from a detailed investigation, only three types of points, Q, R, and S, compete to define the global measure of the workspace performances. Hence, the problem of this Sub-Section is reduced to choosing the worst transmission factor from these points.

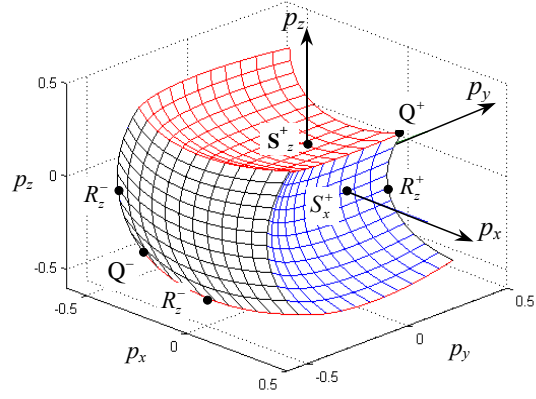

Fig. 8. The $\rho$-bounded workspace and its critical points.

**Vertex points $Q^+$, $Q^-$.** For the points $Q^+$ and $Q^-$, which are located at the intersection of the workspace boundary and the bisector line (i.e. the Q-axis)

$$Q^+: \quad \mathbf{p} = (p_1, p_1, p_1); \quad \boldsymbol{\rho} = (\rho_{max}, \rho_{max}, \rho_{max});$$
$$Q^-: \quad \mathbf{p} = (p_2, p_2, p_2); \quad \boldsymbol{\rho} = (\rho_{min}, \rho_{min}, \rho_{min});$$

the Jacobian is symmetrical, so the transmission factors $\mu_i$ are equal to the inverses $1/\lambda_1$ and $1/\lambda_{2,3}$ of the eigenvalues $\lambda_1 = 1 + 2\chi$ and $\lambda_{2,3} = 1 - \chi$ (see Sub-Section 4.3). The parameter $\chi$ is related to the Cartesian coordinates ($p_x, p_y, p_z = p$) by the expression $\chi = -p / \sqrt{1 - 2p^2}$, where $p_1$, $p_2$ are expressed as

$$p_1 = \frac{1}{3} \left( \rho_{max} - \sqrt{3 - 2\rho_{max}^2} \right); \qquad p_2 = \frac{1}{3} \left( \rho_{min} - \sqrt{3 - 2\rho_{min}^2} \right) \qquad (17)$$

and corresponds to $Q^+$ and $Q^-$ respectively.

**Edge points $R_x^+, ... R_z^-$.** For the points $R_x^+$ and $R_x^-$, which are defined at the intersection of the workspace boundary and the XY-plane

$$R_x^+: \quad \mathbf{p} = (p_1, p_1, 0); \quad \boldsymbol{\rho} = (\rho_{max}, \rho_{max}, \rho_1);$$
$$R_x^-: \quad \mathbf{p} = (p_2, p_2, 0); \quad \boldsymbol{\rho} = (\rho_{min}, \rho_{min}, \rho_2);$$

the inverse Jacobian is

$$\mathbf{J}^{-1}(\chi) = \begin{bmatrix} 1 & \chi & 0 \\ \hline \chi & 1 & 0 \\ \hline \chi/\sqrt{1-\chi^2} & \chi/\sqrt{1-\chi^2} & 1 \end{bmatrix} \qquad (18)$$

where $\chi = -p / \sqrt{1 - p^2}$ and $p$ is equal to either $p_1$ or $p_2$. Using the Orthoglide kinematic equation (1), the Cartesian coordinates may be expressed as

$$p_1 = \left( \rho_{max} - \sqrt{2 - \rho_{max}^2} \right) \Big/ 2; \qquad p_2 = \left( \rho_{min} - \sqrt{2 - \rho_{min}^2} \right) \Big/ 2; \qquad (19)$$

and, subsequently,

$$\rho_1 = \sqrt[4]{\rho_{max}^2 \left(2 - \rho_{max}^2\right)} \in \left(1, \rho_{max}\right); \qquad \rho_2 = \sqrt[4]{\rho_{min}^2 \left(2 - \rho_{min}^2\right)} \in \left(\rho_{min}, 1\right) \qquad (20)$$

Since the matrix (18) is asymmetrical, the velocity transmission factors are to be computed from the product of the Jacobian by its transpose

$$\left(\mathbf{J} \cdot \mathbf{J}^T\right)^{-1} = \begin{bmatrix} 1 + \chi^2 & 2\chi & \left(\chi + \chi^2\right)/\sqrt{1 - \chi^2} \\ 2\chi & 1 + \chi^2 & \left(\chi + \chi^2\right)/\sqrt{1 - \chi^2} \\ \left(\chi + \chi^2\right)/\sqrt{1 - \chi^2} & \left(\chi + \chi^2\right)/\sqrt{1 - \chi^2} & \left(1 + \chi^2\right)/\left(1 - \chi^2\right) \end{bmatrix} \qquad (21)$$

The corresponding characteristic equation may be presented as $\left(\sigma - \left(1 - \chi\right)^2\right) \cdot \left(\sigma^2 - A\sigma + B\right) = 0$, where $A = \left(1 + \chi\right)^2 + \left(1 + \chi^2\right)/\left(1 - \chi^2\right); \quad B = \left(1 + \chi\right)^2$. So, the singular values $\lambda = \sqrt{\sigma}$ are

$$\lambda_1 = 1 - \chi; \qquad \lambda_{2,3} = \sqrt{\frac{1}{2}\left(A \pm \sqrt{A^2 - 4B}\right)} \qquad (22)$$

and the velocity transmission factors can be computed as $1/\lambda_1$, $1/\lambda_2$, and $1/\lambda_3$. It is clear that, because of the symmetry, these assessments are also valid for the remaining points $R_y^+$, $R_y^-$ and $R_z^+$, $R_z^-$.

**Face points $S_x^+, \dots S_z^-$.** For the points $S_x^+$ and $S_x^-$,, which are defined at the intersection of the workspace boundary and the X-axis

$$S_x^+: \quad \mathbf{p} = \left(p_1, 0, 0\right); \; \boldsymbol{\rho} = \left(\rho_{max}, \rho_1, \rho_1\right);$$
$$S_x^-: \quad \mathbf{p} = \left(p_2, 0, 0\right); \; \boldsymbol{\rho} = \left(\rho_{min}, \rho_2, \rho_2\right);$$

the inverse Jacobian is

$$\mathbf{J}\left(\chi\right)^{-1} = \begin{bmatrix} 1 & 0 & 0 \\ \chi & 1 & 0 \\ \chi & 0 & 1 \end{bmatrix} \qquad (23)$$

where $\chi = -p/\sqrt{1 - p^2}$ and $p$ is equal to either $p_1$ or $p_2$. Using the basic kinematic equation (1), the latter may be expressed as $p_1 = \rho_{max} - 1; \quad p_2 = \rho_{min} - 1$, and, subsequently,

$$\rho_1 = \sqrt{\rho_{max}\left(2 - \rho_{max}\right)} \in \left[1, \rho_{max}\right]; \qquad \rho_2 = \sqrt{\rho_{min}\left(2 - \rho_{min}\right)} \in \left[\rho_{min}, 1\right] \qquad (24)$$

Since the matrix (23) is asymmetrical, the velocity amplification factors must be computed from the product of the Jacobian and its transpose

$$\left(\mathbf{J} \cdot \mathbf{J}^T\right)^{-1} = \begin{bmatrix} 2 + \chi^2 & \chi & \chi \\ \chi & 1 & 0 \\ \chi & 0 & 1 \end{bmatrix} \qquad (25)$$

The corresponding characteristic equation may be written as $(\sigma - 1) \cdot (\sigma^2 - 2(1 + \chi^2)\sigma + 1) = 0$.

Hence, the singular values $\lambda = \sqrt{\sigma}$ are

$$\lambda_1 = 1; \quad \lambda_{2,3} = \sqrt{(1 + \chi^2) \pm \chi\sqrt{2 + \chi^2}} \tag{26}$$

and the velocity transmission factors can be computed as $1/\lambda_1$, $1/\lambda_2$, and $1/\lambda_3$. It is clear that similar results are also valid for the points $S_y^+$, $S_y^-$ and $S_z^+$, $S_z^-$.

## 5.2. GLOBAL PERFORMANCE INDICES

After evaluation of the transmission factors at the points Q, R, S, these points can be classified with respect to the influence on the global performance indices $\mu_{\min}$, $\mu_{\max}$ throughout the workspace $W_\rho$ bounded by $\rho \in [\rho_{\min}, \rho_{\max}]$. Here, the global performance indices are defined as the lower and upper bounds of the velocity transmission factors within $W_\rho$. Since the prescribed workspace must be singularity-free, the allowable joint limits should belong to the rectangle $(\rho_{\min}, \rho_{\max}) \in [0, 1] \times [1, \sqrt{1.5}]$.

**Contour plots of global indices**. Detailed investigation of the joint limit rectangle based on both analytical and numerical tools has yielded results presented in Fig. 9, which contains the contour plots of $\mu_{\min}$, $\mu_{\max}$ on the plane $\rho_{\min}$, $\rho_{\max}$. These plots are labelled by the relevant values of the velocity transmission factors and divided in separate areas, which differ by a type of the critical points. For instance, the contour plot for the function $\mu_{\min}(\rho_{\min}, \rho_{\max})$ consists of four areas, $S^-$, $R^-$, $Q^-$ and $Q^+$, where the global transmission factors are defined by the critical points $\{S_x^-, S_y^-, S_z^-\}$, $\{R_x^-, R_y^-, R_z^-\}$, $Q^-$ and $Q^+$ respectively. For comparison purposes, we also show by the dashed/dotted lines the one-dimensional subset of the joint limit rectangle, which corresponds to the "symmetrical" design constrains (i.e. $\mu_{\min} = 1/\mu_{\max}$) imposed either on the full $\rho$-bounded workspace or along the Q-axis only.

As follows from Fig. 9a, the global minimum of the transmission factor can be achieved in either points $S_i^-$, $R_i^-$, or $Q^-$, where $i \in \{x, y, z\}$ and all these indices are equivalent with respect to the $\mu_{\min}$, $\mu_{\max}$ because of the symmetry. It has been proved, that particular expressions for computing of $\mu_{\min}$ are

$$\mu_{\min}(\rho_{\min}, \rho_{\max}) = \begin{cases} \lambda_2 \left(S_i^-\right)^{-1} & \text{for} \quad \rho_{\min} \in \left] 0, \; \rho_{SR} \right] \\ \lambda_2 \left(R_i^-\right)^{-1} & \text{for} \quad \rho_{\min} \in \left] \rho_{SR}, \rho_{RQ} \right] \\ \lambda_2 \left(Q^-\right)^{-1} & \text{for} \quad \rho_{\min} \in \left] \rho_{RQ}, \varphi_{QQ}(\rho_{\max}) \right] \\ \lambda_1 \left(Q^+\right)^{-1} & \text{for} \quad \rho_{\min} \in \left] \varphi_{QQ}(\rho_{\max}), \; 1 \right[ \end{cases} \tag{27}$$

where the subscripts of $\lambda$ define the number of the critical singular value (1...3, in accordance with the above notation), and the critical points $S_i^-$, $R_i^-$ and $Q^-$ are separated by the vertical lines $\rho_{SR} = 0.1093$ and $\rho_{RQ} = 0.2240$ shown in bold in Fig. 9a (corresponding values of the transmission factor are $\mu_{min}(\rho_{SR}) = 0.3232$ and $\mu_{min}(\rho_{RQ}) = 0.4210$). The critical points $Q^+$, $Q^-$ are separated by the curve $\rho_{min} = \varphi_{QQ}(\rho_{max})$, which can be obtained by equating the eigenvalues $\lambda_1 = 1 + 2\chi$ and $\lambda_2 = 1 - \chi$ for $Q^+$ and $Q^-$ respectively. Hence, using the relation between the auxiliary variable $\chi$ and the joint coordinate along the Q-axis $\rho = (1-\chi)/\sqrt{1+2\chi^2}$, the equations for the function $\varphi_{QQ}(\rho_{max})$ can be presented both in the parametric

$$\rho_{min} = (1-\chi)/\sqrt{1+2\chi^2}; \quad \rho_{max} = (1+2\chi)/\sqrt{1+8\chi^2}; \quad \chi \in \, ] \, 0, \, 1/4 \, [ \qquad (28a)$$

and explicit form

$$\rho_{max}^2 = 3(3 - 2\rho_{min}^2)/((9 - 2\rho_{min}^2) - 4\rho_{min}\sqrt{3 - 2\rho_{min}^2}); \quad \rho_{min} \in \, ] \sqrt{0.5}, \, 1.0 \, [. \qquad (28b)$$

It can be also shown that along this curve the corresponding velocity transmission factor $\mu_{min} = 1/(1+2\chi)$ varies from 1 to $2/3$, and the curve is bounded by the points $(1,1)$ and $(\sqrt{0.5}, \sqrt{1.5})$.

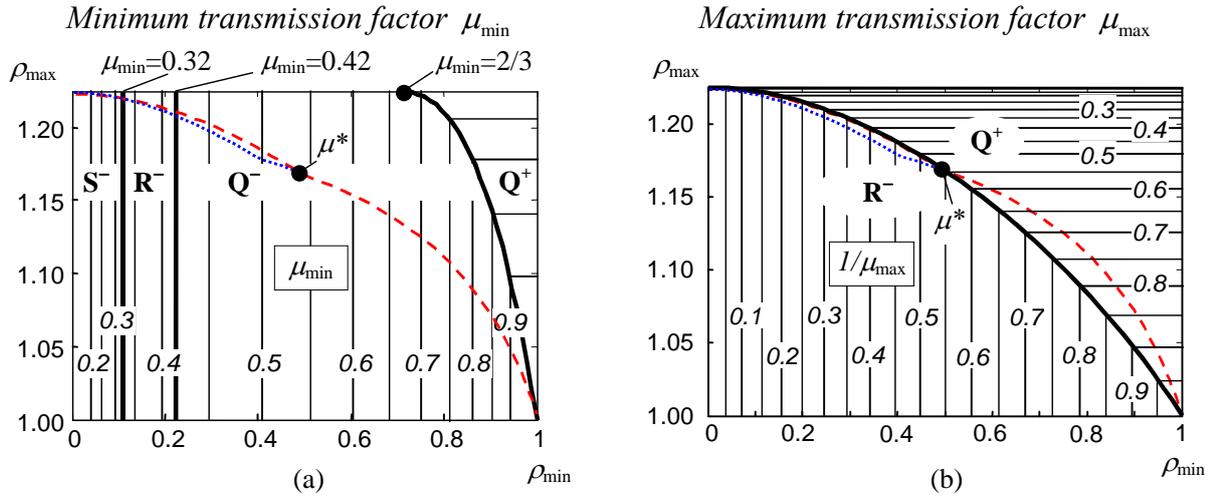

Fig. 9. Contour plots of the global transmission factors $\mu_{min}$, $\mu_{max}$ on the plane $\rho_{min} \times \rho_{max}$ (dashed/ dotted lines correspond to symmetrical constraints for $W_\rho$/ Q-axis; $\mu^* \approx 0.54$).

Similar analysis for the global maximum of $\mu_{max}$ (Fig. 9b) shows that it can be achieved in either point $R_i^-$ or $Q^+$, i.e.

$$\mu_{max}(\rho_{min}, \rho_{max}) = \begin{cases} \lambda_1(R_i^-)^{-1} & \text{for} \quad \rho_{max} \in \, ] \, 1, \, \varphi_{RQ}(\rho_{min}) \, ] \\ \lambda_1(Q^+)^{-1} & \text{for} \quad \rho_{max} \in \, ] \, \varphi_{RQ}(\rho_{min}), \, \sqrt{1.5} \, [ \end{cases} \qquad (29)$$

where the subscripts of $\lambda$ and $R$ have similar meaning as in (27), and the critical points $R_i^-$ and $Q^+$ are separated by the curve $\rho_{\min} = \varphi_{RQ}(\rho_{\max})$. This curve can be obtained by equating the eigenvalues $\lambda_1 = 1 - \chi$ and $\lambda_1 = 1 + 2\chi$ respectively for $R_i^-$ and $Q^+$. Hence, using the relations between the auxiliary variable $\chi$ and the joint coordinates of the Q-points $\rho_Q = (1-\chi)/\sqrt{1+2\chi^2}$ and R-point $\rho_R = (1-\chi)/\sqrt{1+\chi^2}$, the equations for the function $\varphi_{RQ}(\rho_{\max})$ can be presented both in the parametric

$$\rho_{\max} = (1-\chi)/\sqrt{1+2\chi^2}; \qquad \rho_{\min} = (1+2\chi)/\sqrt{1+4\chi^2}; \qquad \chi \in \,]-1/2, 0\,[\,. \qquad (30a)$$

and explicit forms

$$\rho_{\min}^2 = \frac{9(3-2\rho_{\max}^2)}{(15-2\rho_{\max}^2)-4\rho_{\max}\sqrt{3-2\rho_{\max}^2}} \qquad \rho_{\max} \in \,]\,1,\, \sqrt{3/2}\,[\,. \qquad (30b)$$

It can be also shown that along this curve the corresponding velocity transmission factor $\mu_{\max} = 1/(1+2\chi)$ varies from 1 to $\infty$, and the curve is bounded by the points $(1,1)$ and $(0, \sqrt{1.5})$.

**Neighbourhood of the isotropic point**. Since the kinematic design seeks for the quasi-isotropic workspace, it is useful to obtain analytical expressions for $\mu_{\min}$ in the neighbourhood of the isotropic point $(1,1)$, which is completely defined by the Q-axis (see Fig. 9a). As proved above, for both $Q^+$ and $Q^-$ the auxiliary parameter $\chi$ may be expressed via the joint variable as $\chi_Q = \left(1-\rho\sqrt{3-2\rho^2}\right)\!\big/\!\left(2\rho^2-1\right)$, hence in this case (27) may be reduced up to

$$\mu_{\min} = \begin{cases} \dfrac{1}{3} + \dfrac{2\rho_{\min}}{3\sqrt{3-2\rho_{\min}^2}} & \text{if} \quad \rho_{\max} \geq \varphi_{QQ}(\rho_{\max}) \\[3mm] \dfrac{2}{3} + \dfrac{\sqrt{3-2\rho_{\max}^2}}{3\rho_{\max}} & \text{otherwise} \end{cases} \qquad (31)$$

Rewriting these expressions with respect to $\mu_{\min}$ yields

$$Q^-: \;\; \rho_{\min} \geq \frac{3\mu_{\min}-1}{\sqrt{6\mu_{\min}^2-4\mu_{\min}+2}} \;; \qquad\qquad Q^+: \;\; \rho_{\max} \leq \frac{1}{\sqrt{3\mu_{\min}^2-4\mu_{\min}+2}} \;, \qquad (32)$$

which allow to compute the joint limits $\rho_{\min}, \rho_{\max}$ for given design specification $\mu_{\min} \geq \mu^0$ (provided that $\mu_{\min} \geq \mu(\rho_{SR}) \approx 0.32$). For the wider range of $\mu_{\min}$, the relevant equation was solved numerically and corresponding plots are presented in Fig. 10.

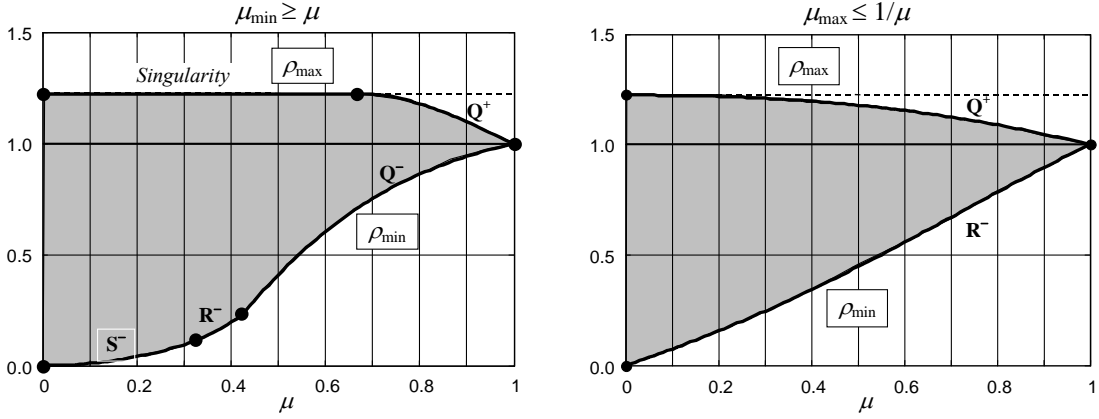

Fig. 10. Computing joint limits for non-symmetrical constraints
(independent upper and lower bounding)

Similar expressions for $\mu_{max}$ can be derived by substitution of

$$\chi_R = \left(1 - \rho_{min}\sqrt{2 - \rho_{min}^2}\right)\bigg/\left(\rho_{min}^2 - 1\right); \qquad \chi_Q = \left(1 - \rho_{max}\sqrt{3 - 2\rho_{max}^2}\right)\bigg/\left(2\rho_{max}^2 - 1\right)$$

into equations for the roots $\lambda_1 = 1 - \chi_R$ and $\lambda_1 = 1 + 2\chi_Q$ and relevant simplification:

$$\mu_{max} = \begin{cases} \dfrac{1}{3} + \dfrac{2\rho_{max}}{3\sqrt{3 - 2\rho_{max}^2}} & \text{if} \quad \rho_{min} \le \varphi_{RQ}(\rho_{max}) \\[3mm] \dfrac{1}{2} + \dfrac{\sqrt{2 - \rho_{min}^2}}{2\rho_{min}} & \text{otherwise} \end{cases} \tag{33}$$

Rewriting these expressions with respect to $\mu_{max}$ yields

$$\text{R}^-: \quad \rho_{min} \ge \frac{1}{\sqrt{2\mu_{max}^2 - 2\mu_{max} + 1}} \; ; \qquad \text{Q}^+: \quad \rho_{max} \le \frac{3\mu_{max} - 1}{\sqrt{6\mu_{max}^2 - 4\mu_{max} + 2}} \; , \tag{34}$$

which make it possible to compute the joint limits $\rho_{min}, \rho_{max}$ for given design specification $\mu_{max} \le 1/\mu$ (see Fig. 10).

**Symmetrical design specification.** The above results can be also used in the case of the "symmetrical" design specification, which assumes the inverse relations between the upper and lower bound on the transmission factor ($\mu_{min} = 1/\mu_{max}$), when the feasible workspace is defined by the expression

$$\min_{\mathbf{p} \in W_\rho} \left\{\mu_{min}(\mathbf{p}), \; 1/\mu_{max}(\mathbf{p})\right\} \ge \mu \tag{35}$$

in which $\mu \in [0, 1]$ is the value of the prescribed performance measure (for instance, it is natural to set the transmission factor to be in the range $[1/2, 2]$ or $[1/3, 3]$, as it was proposed in [16]). As follows from the related analysis, in this case only two combinations of critical points are possible: $(\text{Q}^+, \text{Q}^-)$ if $\mu \ge \mu^*$ and $(\text{Q}^+, R_i^-)$ if $\mu < \mu^*$, where $\mu^* \approx 0.5387$, $\rho_{min}(\mu^*) \approx 0.4892$, $\rho_{max}(\mu^*) \approx 1.1700$ (see Fig. 9). So, as follows from (31) and (33), the joint limits for the symmetrical design specification can be computed as

$$\mathbf{Q}^+: \quad \rho_{max} \leq \frac{3-\mu}{\sqrt{2\mu^2 - 4\mu + 6}}, \qquad 0 < \mu < 1 \tag{36a}$$

$$\mathbf{Q}^-: \quad \rho_{min} \geq \frac{3\mu - 1}{\sqrt{6\mu^2 - 4\mu + 2}}, \quad \mu^* \leq \mu < 1; \qquad \mathbf{R}^-: \quad \rho_{min} \geq \frac{\mu}{\sqrt{\mu^2 - 2\mu + 2}}, \quad 0 < \mu < \mu^* \tag{36b}$$

This relation is shown in Fig. 11a on the $\rho_{min} \times \rho_{max}$ plane and also plotted in Fig. 11b against the velocity transmission factor $\mu$. They make it possible to evaluate the global manipulator performance for the whole $\rho$-bounded workspace $W_\rho$ (for given joint limits) or to compute the joint limits that guarantee the desired performances throughout $W_\rho$.

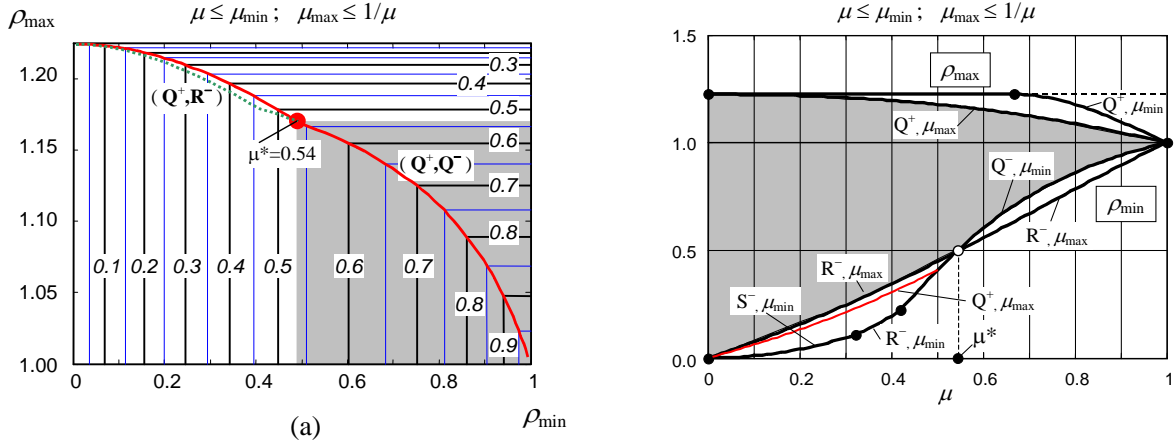

(a)

Fig. 11. Transmission factors (a) and joint limits (b) for symmetrical constraints within $W_\rho$ (dashed line shows symmetrical constraints for Q-axis only).

## 5.3. DEFINING A CUBIC WORKSPACE

Since the prescribed Cartesian workspace has a cubic shape, let us first define the largest cube that ensures the desired transmission factors $[\mu, 1/\mu]$ through it, while temporarily releasing the joint limits constraints. It is obvious, that due to the Orthoglide symmetrical architecture, the cube faces must be parallel to the *xy*, *xz* and *yz* planes. So, the constraint (15) may be rewritten as

$$\min_{\mathbf{p} \in W_p} \{\mu_{min}(\mathbf{p}), \ 1/\mu_{max}(\mathbf{p})\} \geq \mu \tag{37}$$

where $W_p$ denotes the *p*-bounded workspace determined by $p_x, p_y, p_z \in [p_{min}, p_{max}]$. Applying the above notation, the cube may be also defined by its two opposite vertices $\mathbf{Q}^+, \mathbf{Q}^-$ located on the bisector line. Detailed investigation of $W_p$ using expressions from previous Sub-Sections is summarised in the following statement.

**Proposition 1**. *If the prescribed symmetrical bounds* $\mu(\mathbf{p}) \in [\mu, 1/\mu]$, $0 < \mu < 1$ *on the velocity transmission factors are satisfied at the* Q-*axis points* $\mathbf{Q}^+, \mathbf{Q}^-$, *then these bounds are satisfied throughout the cubic workspace* $W_p$ *defined by the vertices* $\mathbf{Q}^+, \mathbf{Q}^-$ *(and vice versa).*

The proof of the proposition uses convexity of the workspace hull bounded by $\mu(\mathbf{p}) \in [\mu, 1/\mu]$ and is based on comparing the Jacobian singular values in the critical points of the cubic workspace and on the expressions for the joint limits

$$\rho(Q^+) = \frac{3-\mu}{\sqrt{2\mu^2 - 4\mu + 6}}; \quad \rho(Q^-) = \max\left\{\frac{3\mu-1}{\sqrt{6\mu^2 - 4\mu + 2}}; \frac{\mu}{\sqrt{2\mu^2 - 4\mu + 3}}\right\}, \quad (38)$$

which are derived from (31), (33). It should be noted that the $\rho$-bounded workspace defined by the same vertices $Q^+, Q^-$ does not satisfy the bounds $[\mu, 1/\mu]$ if $\mu < \mu^* \approx 0.54$ (see Sub-Section 5.2), but the $p$-bounds remove the critical points $R_x^+, R_y^+, \ldots R_z^-$. Besides, utilising the cubic workspace with the vertices $Q^+, Q^-$ requires a certain enlargement of the upper joint limit. Really, as follows from the basic Orthoglide equations (1), the $p$-bounded Cartesian workspace maps into the joint space portion, which is contained in the parallelepiped

$$\rho_x, \rho_y, \rho_z \in \left[\rho(Q^-), \ p(Q^+) + 1\right], \quad (39)$$

where $\rho(Q^-) = p_{min} + \sqrt{1 - 2p_{min}^2}$ is the joint coordinate at $Q^-$, $p(Q^+) = p_{max}$ is the Cartesian coordinate at $Q^+$, and $p_{min}, p_{max}$ depend on the desired transmission factor bound $\mu$ and are computed from (31), (33). It can also be easily proved that $p(Q^+) + 1 > \rho(Q^+)$, since $\rho(Q^+) = p_{max} + \sqrt{1 - 2p_{max}^2}$ (see Sub-Section 5.1). Hence, this increase in the upper joint limit may lead to the singularities included in the corresponding $\rho$-bounded workspace, which can be avoided only by adding some "software joint limits" (based on verifying of inequalities more complicated than $\rho_{min} \leq \rho_x \leq \rho_{max}$ and similar). Relevant computations showed that the singularity problem arises for the transmission factors $\mu \leq 1 - \sqrt{\sqrt{1.5} - 1} \approx 0.53$ for which $p(Q^+) + 1 \geq \sqrt{1.5}$.

The alternative approach for defining the cubic workspace assumes that the joint limits corresponding to $Q^+, Q^-$ cannot be violated. Hence, the desired cube $W_p^o$ should be completely enclosed in the $\rho$-bounded space $W_\rho$. It is obvious that $W_p^o \subset W_p$, so within the cube $W_p^o$ the manipulator possesses the desired kinematic properties. Dimension of the cube $W_p^o$ and its spatial location are defined by the following proposition.

**Proposition 2**. *If the prescribed symmetrical bounds $\mu(\mathbf{p}) \in [\mu, 1/\mu]$, $0 < \mu < 1$ are satisfied in the Q-axis points $Q^+, Q^-$, then the largest cube enclosed in the $\rho$-bounded workspace $W_\rho$ is defined by the vertices $Q_*^+$ and $Q^-$, where $Q_*^+ \in \left[Q^+Q^-\right]$ and $p(Q_*^+) = \rho(Q^+) - 1$.*

The proof of this proposition is based on comparing the Cartesian coordinates for the critical points Q, R, S (see Fig. 8) and also uses the expression for the upper joint limit from Proposition 1. It is clear, that in this case the $\rho$-bounded workspace defined by $Q^+, Q^-$ is singularity-free, but its global performances may be out of the design specifications in $R_2$-points and their neighbourhood if $\mu < 0.54$ (see Sub-Section 5.2). So, the designer may choose the third design strategy, which guarantees satisfaction of the design specification both in the $\rho$-bounded and $p$-bounded workspaces $W_\rho$ and $W_p^o$. The latter is based on the following corollary combining results from Propositions 1 and 2.

**Corollary**. *If the prescribed symmetrical bounds* $\mu(\mathbf{p}) \in [\mu, 1/\mu]$, $0 < \mu < 1$ *are satisfied within the* $\rho$-*bounded workspace defined by the vertices* $Q^+, Q^-$, *then they are also satisfied within the* $p$-*bounded workspace defined by the vertices* $Q_*^+$ *and* $Q^-$, *with* $p(Q_*^+) = \rho(Q^+) - 1$.

These Propositions and Corollary give the designer three different methods ("design strategies") for computing the joint limits and dextrous Cartesian workspace of the normalised manipulator ($L$=1), which afterwards must be scaled to achieve the prescribed workspace size. The methods are summarised in Table 3 and yield three Pareto-optimal solutions with respect to the design goals stated in Section 2. As follows from the propositions, all strategies ensure satisfaction of the design specification within the prescribed cubic workspace $W_p$, but differ by the manipulator performances in the remaining part $W_\rho \setminus W_p$. It should be noted, that the primary version of the first method, for $\mu \geq \mu^*$, was developed in the previous paper [16], while here it is generalised for the full range of the transmission factor.

Table 3

Computing joint limits for the unit Orthoglide

(dots on the Q-axis show location of $Q^+, Q^-$ for different design strategies).

| Design strategies | Remarks |
|---|---|
| **Design strategy #1**<br><br>(i) Compute points $Q^+, Q^-$ to achieve required transmission factors along segment $Q^+Q^-$.<br><br>(ii) Locate the cube vertices in points $Q^+, Q^-$ to define the cubic workspace $W_p$.<br><br>(iii) Adjust the joint limits to include the $p$-bounded workspace $W_p$ inside the $\rho$-bounded one. | 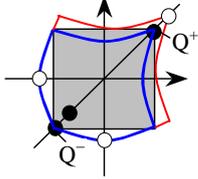<br><br>$\rho_{\min} = \rho_{Q^-}$ ; $\rho_{\max} = 1 + p_{Q^+}$ ;<br>$p_{\min} = p_{Q^-}$ ; $p_{\max} = p_{Q^+}$ ;<br><br>Inside the cube, design specifications are satisfied, but outside it, they are violated and even singularities exist<br><br>if $\mu \leq 0.53$ |
| **Design strategy #2**<br><br>(i) Compute points $Q^+, Q^-$ to achieve required transmission factor along segment $Q^+Q^-$ and set joint limits according to these points.<br><br>(ii) Inscribe the cube inside the $\rho$-bounded workspace $W_\rho$ to define the cubic workspace $W_p$. | 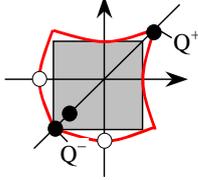<br><br>$\rho_{\min} = \rho_{Q^-}$ ; $\rho_{\max} = \rho_{Q^+}$ ;<br>$p_{\min} = p_{Q^-}$ ; $p_{\max} = \rho_{Q^+} - 1$ ;<br><br>Workspace is singularity-free but, outside the cube, performances are out of design specifications<br><br>if $\mu < 0.54$ |
| **Design strategy #3**<br><br>(i) Compute points $Q^+, Q^-$ to achieve required transmission factor within the $\rho$-bounded workspace $W_\rho$ and set according joint limits.<br><br>(ii) Inscribe the cube inside the $\rho$-bounded workspace $W_\rho$ to define the cubic workspace $W_p$. | 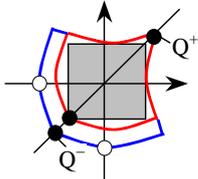<br><br>$\rho_{\min} = \rho_{Q^-}$ ; $\rho_{\max} = \rho_{Q^+}$ ;<br>$p_{\min} = p_{Q^-}$ ; $p_{\max} = \rho_{Q^+} - 1$ ;<br><br>Both $\rho$- and $p$-bounded workspaces are singularity-free and meet design specifications |

It is obvious that correctness of the above statements for the transmission factor guarantees their correctness for the manipulability and condition number indices, which may be directly expressed via the singular values. Also, for real-life problems, the designer can prefer one of the solutions to other ones taking into account a number of additional engineering constraints and objectives, which cannot be implicitly expressed in the frames of the model used in this paper.

## 6.    NUMERICAL EXAMPLES AND DISCUSSIONS

To compare the proposed design approaches, let us apply them to the design of the Orthoglide for the unit Cartesian workspace $c \times c \times c$, $c = 1$ with the transmission factors bounds $0.5 \leq \mu \leq 2.0$. As stated above, the design process includes two main stages: (i) defining the joint limits $\rho_{min}, \rho_{max}$ for the normalised manipulator with the link length $L = 1$, and (ii) scaling the manipulator parameters $\left( \rho_{min}, \rho_{max}, p_{min}, p_{max}, L \right)$ to achieve the prescribed workspace size.

For the normalised manipulator, the Q-axis technique gives the following ranges for the joint/Cartesian coordinates and corresponding transmission factors within the $\rho$- and $p$-bounded workspaces $\mu\left(W_\rho\right)$, $\mu\left(W_p\right)$, which are called below J-Space and C-Space, respectively (i.e. "Joint-coordinate bounded space" and "Cartesian-coordinate bounded space"):

J-Space$_0$:     $\rho \in \left[\ 0.4082,\ 1.1785\right]$;     $\Delta\rho = 0.7703$;     $\mu\left(W_p\right) \in [0.50,\ 2.16]$

C-Space$_0$:     $p \in \left[-0.4082,\ 0.2357\right]$;     $\Delta p = 0.6440$;     $\mu\left(W_p\right) \in [0.50,\ 2.00]$

*Strategy* #1 assumes that the cube with the edge $\Delta p$ is used as the prescribed Cartesian workspace. It requires increasing the upper joint limit to make all points of the cube attainable. According to Sub-Section 5.3, the enlarged joint space is defined as

J-Space$_1$:     $\rho \in \left[\ 0.4082,\ 1.2357\right]$;     $\Delta\rho = 0.8275$;     $\mu\left(W_\rho\right) = [0.50,\ \infty[$

So, since $\rho_{max} > \sqrt{1.5}$, the obtained $\rho$-bounded workspace includes parallel singularities, which may be eliminated by additional software constrains on the joint coordinates. For instance, the inequality $\rho_x + \rho_y + \rho_z \leq 3\rho_{Q^+}$ removes the singularities from $W_\rho$ and restores the original transmission factors $[0.50,\ 2.16]$.

*Strategy* #2 keeps the original $\rho$-bounded workspace, within which is located the prescribed cube. Computing relevant parameters gives:

C-Space$_2$:     $p \in \left[-0.4082,\ 0.1785\right]$;     $\Delta p = 0.5868$;     $\mu\left(W_p\right) = [0.50, 2.00]$

In this case, the cube is smaller but the workspace is singularity-free and possesses reasonable kinematic properties both inside and outside the cube.

*Strategy* #3 provides the desired transmission factors for the whole $\rho$-bounded workspace, which is computed in accordance with Table 3 and is defined as

J-Space$_3$:     $\rho \in \left[\ 0.4472,\ 1.1785\right]$;     $\Delta\rho = 0.7313$;     $\mu\left(W_\rho\right) = \left[0.52,\ 2.00\right]$

Then, inscribing the largest cube gives

C-Space$_3$:     $p \in \left[-0.3884,\ 0.1785\right]$;     $\Delta p = 0.5669$;     $\mu\left(W_p\right) = \left[0.52,\ 1.87\right]$

that overtakes the required kinematic performances for the cube but ensures them for the whole $\rho$-bounded workspace.

After defining normalised manipulator parameters, the obtained cubic workspaces must be adjusted to the prescribed size $c \times c \times c$ by scaling the manipulator dimensions and joint/Cartesian coordinates. Computing the scaling factor $\eta = c/\Delta\rho$ for $c = 1$ yields 1.553, 1.704 and 1.764 for the first, second and third design strategies respectively. Design results after scaling are summarised in Table 4, which also contains ratio of the actuated joints range $\Delta\rho$ to the cube size $c$. These results shows that all obtained solutions are Pareto-optimal with respect to the vector criterion $\left(L, \mu_{\min}, \mu_{\max}\right)$ with the goals $L \rightarrow \min$ and $\mu_{\min}, \mu_{\max} \rightarrow 1.0$. Indeed, Strategy #1 yields the smallest Orthoglide dimensions (about 12% less than the third one), but the worst kinematic properties outside the cube (with singularities). In contrast, Strategy #3 guarantees the best kinematic performances for the price of the largest manipulator links, while Strategy #2 gives an intermediate solution ensuring the compromise between the link length and transmission factors. Hence, none of the strategies can be given a preference within the frames of the kinematic model and, in real-life applications, all these solutions should be presented to the designer who may evaluate them by taking into account a number of additional technical constraints and goals.

Table 4

Orthoglide parameters and performances for $W_p = 1 \times 1 \times 1$ and $0.5 \leq \mu \leq 2.0$

| Design strategy | $L$ | $\rho_{\min}$ | $\rho_{\max}$ | $\Delta\rho$ | $c/\Delta\rho$ | $\mu\left(W_p\right)$ | $\mu\left(W_\rho\right)$ |
|---|---|---|---|---|---|---|---|
| #1 | 1.553 | 0.634 | 1.919 | 1.285 | 0.7782 | 0.500 … 2.000 | singularity |
| #2 | 1.704 | 0.696 | 2.009 | 1.313 | 0.7618 | 0.500 … 2.000 | 0.500 … 2.158 |
| #3 | 1.764 | 0.789 | 2.079 | 1.290 | 0.7752 | 0.518 … 1.869 | 0.518 … 2.000 |

Finally, let us demonstrate application of the proposed technique to the design of the Orthoglide prototype, which has been built in Institut de Recherche en Communications et Cybernétique de Nantes (IRCCyN). The prescribed performances of the manipulator are: Cartesian velocity and acceleration in the isotropic point 1.2 $m/s$ and 14 $m/s^2$ ; payload 4 $kg$ ; cubic Cartesian workspace size $200 \times 200 \times 200$ $mm$ ; transmission factor range $0.5 \div 2.0$ . Application of the design strategies #1,…3 yielded the Orthoglide link lengths 310.6, 340.9, 352.8 $mm$ respectively. Taking into account additional technical goals related to the manipulator mass and dynamic performances, the preference was given to the solution with the smallest link length. Corresponding joint limits are $\rho_{min} = 126.8$ $mm$ and $\rho_{max} = 383.8$ $mm$. To remove singularities, the software constraint $\rho_x + \rho_y + \rho_z \leq 3\rho_{Q^*}$ were used where $3\rho_{Q^*} = 1098.1$ $mm$. As follows from simulation and laboratory experiments, the prototype ensures required transmission factors within the prescribed cubic workspace [0.50, 2.00] and also their reasonable values outside the cube [0.50, 2.16]. However, during tuning of the control system, it was noticed rather high sensitivity of the kinematic performances with respect to the joint encoder offset. For instance, the 5 $mm$ offset leads to changing of the Cartesian cube transmission factors to [0.50, 2.42]. The 10 $mm$ offset increases their range up to [0.50, 3.42]. This imposes strict requirements on the assembly accuracy and motivates dedicated research on the Orthoglide calibration.

## 7.    CONCLUSIONS

The paper focuse on the parametrical synthesis of the Orthoglide, a parallel manipulator for 3-axis rapid machining applications, which combines advantages of both serial mechanisms (regular workspace and homogeneous performances) and parallel kinematic architectures (good dynamic performances). Three strategies have been proposed to define the Orthoglide geometric parameters as functions of a cubic workspace size and dextrous properties expressed by bounds on the velocity transmission factors, manipulability or the Jacobian condition number. Low inertia and intrinsic stiffness have been set as additional design goals expressed by the minimal link length requirement.

In contrast to previous works, we proposed several Pareto-optimal solutions of the design problem, which differ by the manipulator performances outside the prescribed Cartesian cube (but within the workspace bounded by the actuated joint limits). Taking into account linear relation between the manipulator parameters and the cubic workspace size, the design process is decomposed in two stages: (i) defining the actuated joint limits and the largest cube size/location to satisfy the dexterity goals for the normalised manipulator; (ii) scaling the normalised manipulator to satisfy a specification on the cubic workspace size.

For each design strategy, we proposed analytical expressions for computing the Orthoglide parameters, which were based on the "critical points" concept that allows evaluating the global performance indices through the joint-bounded or cubic workspaces without their exhaustive exploration. We also proved two propositions describing relations between these workspace sizes and kinematic performances within them. It was shown, that independently of the applied strategy, the workspace includes the fully-isotropic point where any linear joint displacement yields similar manipulator tool displacement, like in a serial XYZ-machine. So, the synthesis is aimed at specifying the cubic volume around this point, which meets the dexterity goals. The related design parameters are the actuated joint limits and manipulator link lengths.

The proposed design strategies have been illustrated by numerical examples with the dexterity specification expressed by the velocity transmission factor. We obtained three Pareto-optimal solutions ensuring the required kinematic properties within the cubic workspace but providing wider range of the transmission factor outside the cube (this range is decreased monotonously while the manipulator link length is increased). Hence, no one of the strategies can be given a preference within the frames of the kinematic model and, in real-life applications, all the solutions should be presented to the designer who should evaluate them taking into account additional technical constraints and goals.

The developed technique has been also applied to the design of the Orthoglide prototype, which has been successfully built and tested in IRCCyN (Nantes, France). However, experiments with this manipulator showed rather high sensitivity of the kinematic performances with respect to the joint encoder offsets, which motivates further research on the Orthoglide calibration.

## REFERENCES

[1]    J.-P. Merlet, Parallel Robots, Kluwer Academic Publishers, Dordrecht, 2000.

[2]    J. Tlusty, J.C. Ziegert, S. Ridgeway, Fundamental Comparison of the Use of Serial and Parallel Kinematics for Machine Tools, CIRP Annals 48 (1) (1999) 351-356.

[3]    P. Wenger, C. Gosselin, B. Maille, A comparative study of serial and parallel mechanism topologies for machine tools, in: Proceedings of PKM'99, Milan, Italy, 1999, pp. 23–32.

[4]    L.W. Tsai, Robot Analysis: the Mechanics of Serial and Parallel Manipulators; John Wiley & Sons, New York, 1999.

[5]    I. Bonev, The Parallel Mechanism Information Center (http:// www.parallemic.org).

[6]   J. Kim, C. Park, Performance analysis of parallel manipulator architectures for CNC machining applications, in: Proceedings of the IMECE Symposium on Machine Tools, Dallas, TX, 1997.

[7]   Ph. Wenger, C. Gosselin D. Chablat, A Comparative Study of Parallel Kinematic Architectures for Machining Applications, in: Proceedings of the 2nd Workshop on Computational Kinematics, Seoul, Korea, 2001, pp. 249-258.

[8]   F. Rehsteiner, R. Neugebauer, S. Spiewak, F. Wieland, Putting parallel kinematics machines (PKM) to productive work, CIRP Annals, 48 (1) (1999) 345–350.

[9]   C.-M. Luh, F.A. Adkins, E.J. Haug, C.C. Qui, Working capability analysis of Stewart platforms, Journal of Mechanical Design 118 (6) (1996) 89-91.

[10]  J.-P. Merlet, Determination of 6D workspace of Gough-type parallel manipulator and comparison between different geometries, International Journal of Robotic Research, 19 (9) (1999) 902–916.

[11]  M. Carricato, V. Parenti-Castelli, Singularity-Free Fully-Isotropic Translational Parallel Mechanisms, International Journal of Robotics Research 21 (2) (2002) 161-174.

[12]  X. Kong, C.M. Gosselin, A Class of 3-DOF Translational Parallel Manipulators with Linear I-O Equations, In: Workshop on Fundamental Issues and Future Research Directions for Parallel Mechanisms and Manipulators, Quebec, Canada, 2002, pp. 161-174.

[13]  H. S. Kim, L.W. Tsai, Design Optimization of a Cartesian Parallel Manipulator, Journal of Mechanical Design 125 (1) (2003) 43-51.

[14]  P. Wenger, D. Chablat, Kinematic Analysis of a New Parallel Machine Tool: the Orthoglide, in: Advances in Robot Kinematic, J. Lenar¡ci¡c and M. M. Stani¡sic´, Eds., Norwell, MA, Kluwer Academic Publishers, 2000, pp. 305–314.

[15]  D. Chablat, P. Wenger, J.P. Merlet, Workspace Analysis of the Orthoglide Using Interval Analysis, in: Advances in Robot Kinematic, J. Lenar¡ci¡c and F. Thomas, Eds., Norwell, MA, Kluwer Academic Publishers, 2002, pp. 397–406.

[16]  D. Chablat, Ph. Wenger, Architecture Optimization of a 3-DOF Parallel Mechanism for Machining Applications, the Orthoglide, IEEE Transactions On Robotics and Automation 19 (3) (2003), pp. 403-410.

[17]  A.Pashkevich, D. Chablat, P. Wenger, Kinematics and Workspace Analysis of a Three-Axis Parallel Manipulator: the Orthoglide, to appear in Robotica 2004.)

[18]  K.E. Zanganeh, J. Angeles, Kinematic isotropy and the optimum design of parallel manipulators, International Journal of Robotic Research 16 (2) (1997) 185–197.

[19]  T. Huang, D. Whitehouse, Local dexterity, optimal architecture and optimal design of parallel machine tools, CIRP Annals 47 (1) (1998) 347–351.

[20]  X-J. Liu, Z-L. Jin, F. Gao, Optimum design of 3-DOF spherical parallel manipulators with respect to the conditioning and stiffness indices, Mechanism and Machine Theory 35 (9) (2000) 1257-1267.

[21]  T.Yoshikawa, Manipulability of robot mechanisms, International Journal of Robotic Research 4 (2) (1985) 3–9.